%% file: paper.tex
\documentclass[runningheads,a4paper]{llncs}

\usepackage{amssymb}
\usepackage{amsmath}
\usepackage{graphics, graphicx}
\graphicspath{ {./images/} }

\usepackage[table,pdftex]{xcolor}
\usepackage{xspace}
\usepackage[colorlinks=true,citecolor=blue,urlcolor=black]{hyperref}
\usepackage{subfig}
\usepackage[colorinlistoftodos]{todonotes}
\usepackage{multirow, tabularx}

\newcolumntype{C}{@{\extracolsep{0.2cm}}c@{\extracolsep{0pt}}}%
\usepackage{caption}
\begin{document}

\mainmatter  
\title{Real-time Prediction of Segmentation Quality}
\titlerunning{Real-time Prediction of Segmentation Quality}

\author{R. Robinson$^1$, O. Oktay$^1$, W. Bai$^1$, V.V. Valindria$^1$, M.M. Sanghvi$^{3,4}$, N. Aung$^{3,4}$, J.M. Paiva$^3$, F. Zemrak$^{3,4}$, K. Fung$^{3,4}$, E. Lukaschuk$^5$, A.M. Lee$^{3,4}$, V. Carapella$^5$, Y.J. Kim$^{5,6}$, B. Kainz$^1$, S.K. Piechnik$^5$, S. Neubauer$^5$, S.E. Petersen$^{3,4}$, C. Page$^2$, D. Rueckert$^1$, B. Glocker$^1$}

\authorrunning{Robinson et al.}


\institute{
$^1$ BioMedIA Group, Deptartment of Computing, Imperial College London, UK\\
$^2$ Research \& Development, GlaxoSmithKline, UK\\
$^3$ NIHR Barts Biomedical Research Centre, Queen Mary University London, UK\\
$^4$ Barts Heart Centre, Barts Health NHS Trust, London, UK\\
$^5$ Radcliffe Department of Medicine, University of Oxford, UK\\
$^6$ Severance Hospital, Yonsei University College of Medicine, South Korea\\
}

\maketitle

\input{sections/abstract}
\input{sections/introduction}
\input{sections/method}
\input{sections/results}
\input{sections/conclusion}

\paragraph{\textbf{Acknowledgements:}} RR is funded by KCL\&Imperial EPSRC CDT in Medical Imaging (EP/L015226/1) and GlaxoSmithKline; VV by Indonesia Endowment for Education (LPDP) Indonesian Presidential PhD Scholarship; KF supported by The Medical College of Saint Bartholomew’s Hospital Trust. AL and SEP acknowledge support from NIHR Barts Biomedical Research Centre and EPSRC program grant (EP/P001009/ 1). SN and SKP are supported by the Oxford NIHR BRC and the Oxford British Heart Foundation Centre of Research Excellence. This project supported by the MRC (grant number MR/L016311/1). NA is supported by a Wellcome Trust Research Training Fellowship (203553/Z/Z). The authors SEP, SN and SKP acknowledge the British Heart Foundation (BHF) (PG/14/89/31194). BG received funding from the ERC under Horizon 2020 (grant agreement No 757173, project MIRA, ERC-2017-STG).

\bibliographystyle{splncs}
\bibliography{cites}

\end{document}

%% file: sections/abstract.tex

\begin{abstract}

Recent advances in deep learning based image segmentation methods have enabled real-time performance with human-level accuracy. However, occasionally even the best method fails due to low image quality, artifacts or unexpected behaviour of black box algorithms. Being able to predict segmentation quality in the absence of ground truth is of paramount importance in clinical practice, but also in large-scale studies to avoid the inclusion of invalid data in subsequent analysis.

In this work, we propose two approaches of real-time automated quality control for cardiovascular MR segmentations using deep learning. First, we train a neural network on 12,880 samples to predict Dice Similarity Coefficients (DSC) on a per-case basis. We report a mean average error (MAE) of 0.03 on 1,610 test samples and 97\% binary classification accuracy for separating low and high quality segmentations. Secondly, in the scenario where no manually annotated data is available, we train a network to predict DSC scores from estimated quality obtained via a reverse testing strategy. We report an $\mathrm{MAE} = 0.14$ and 91\% binary classification accuracy for this case. Predictions are obtained in real-time which, when combined with real-time segmentation methods, enables instant feedback on whether an acquired scan is analysable while the patient is still in the scanner. This further enables new applications of optimising image acquisition towards best possible analysis results.

\end{abstract}

%% file: sections/introduction.tex

\section{Introduction}

Finding out that an acquired medical image is not usable for the intended purpose is not only costly but can be critical if image-derived quantitative measures should have supported clinical decisions in diagnosis and treatment. Real-time assessment of the downstream analysis task, such as image segmentation, is highly desired. Ideally, such an assessment could be performed while the patient is still in the scanner, so that in the case an image is not analysable, a new scan could be obtained immediately (even automatically). Such a real-time assessment requires two components, a real-time analysis method and a real-time prediction of the quality of the analysis result. This paper proposes a solution to the latter with a particular focus on image segmentation as the analysis task.

Recent advances in deep learning based image segmentation have brought highly efficient and accurate methods, most of which are based on Convolutional Neural Networks (CNNs). However, even the best method will occasionally fail due to insufficient image quality (e,g., noise, artefacts, corruption) or show unexpected behaviour on new data. In clinical settings, it is of paramount importance to be able to detect such failure cases on a per-case basis. In clinical research, such as population studies, it is important to be able to detect failure cases in automated pipelines, so invalid data can be discarded in the subsequent statistical analysis.

Here, we focus on automatic quality control of image segmentation. Specifically, we assess the quality of automatically generated segmentations of cardiovascular MR (CMR) from the UK Biobank (UKBB) Imaging Study \cite{Petersen2017}.

Automated quality control is dominated by research in the natural-image domain and is often referred to as image quality assessment (IQA). The literature proposes methodologies to quantify the technical characteristics of an image, such as the amount of blur, and more recently a way to assess the aesthetic quality of such images \cite{Bosse2016}. In the medical image domain, IQA is an important topic of research in the fields of image acquisition and reconstruction. An example is the work by Farzi et al. \cite{Farzi2016} proposing an unsupervised approach to detect artefacts. Where research is conducted into the quality or accuracy of image segmentations, it is almost entirely assumed that there is a manually annotated ground truth (GT) labelmap available for comparison. Our domain has seen little work on assessing the quality of generated segmentations particularly on a per-case basis and in the absence of GT.

\paragraph{\textbf{Related Work:}} Some previous studies have attempted to deliver quality estimates of automatically generated segmentations when GT is unavailable. Most methods tend to rely on a reverse-testing strategy. Both Reverse Validation \cite{Zhong2010} and Reverse Testing \cite{Fan2006} employ a form of cross-validation by training segmentation models on a dataset that are then evaluated either on a different fold of the data or a separate test-set. Both of these methods require a fully-labeled set of data for use in training. Additionally, these methods are limited to conclusions about the quality of the segmentation algorithms rather than the individual labelmaps as the same data is used for training and testing purposes.

Where work has been done in assessing individual segmentations, it often also requires large sets of labeled training data. In \cite{Kohlberger2012a} a model was trained using numerous statistical and energy measures from segmentation algorithms. Although this model is able to give individual predictions of accuracy for a given segmentation, it again requires the use of a fully-annotated dataset. Moving away from this limitation, \cite{Valindriaa,Robinson2017} have shown that applying Reverse Classification Accuracy (RCA) gives accurate predictions of traditional quality metrics on a per-case basis. They accomplish this by comparing a set of reference images with manual segmentations to the test-segmentation, evaluating a quality metric between these, and then taking the best value as a prediction for segmentation quality. This is done using a set of only 100 reference images with verified labelmaps. However, the time taken to complete RCA on a single segmentation is prohibits real-time quality control frameworks: around 11 minutes.

\paragraph{\textbf{Contributions:}} In this study, we show that applying a modern deep learning approach to the problem of automated quality control in deployed image-segmentation frameworks can decrease the per-case analysis time to the order of milliseconds whilst maintaining good accuracy. We predict Dice Similarity Coefficient (DSC) at large-scale analyzing over 16,000 segmentations of images from the UKBB. We also show that measures derived from RCA can be used to inform our network removing the need for a large, manually-annotated dataset. When pairing our proposed real-time quality assessment with real-time segmentation methods, one can envision new avenues of optimising image acquisition automatically toward best possible analysis results.

%% file: sections/method.tex

\section{Method \& Material}
\label{sec:method}

\begin{figure}[t]
	\includegraphics[width=\linewidth]{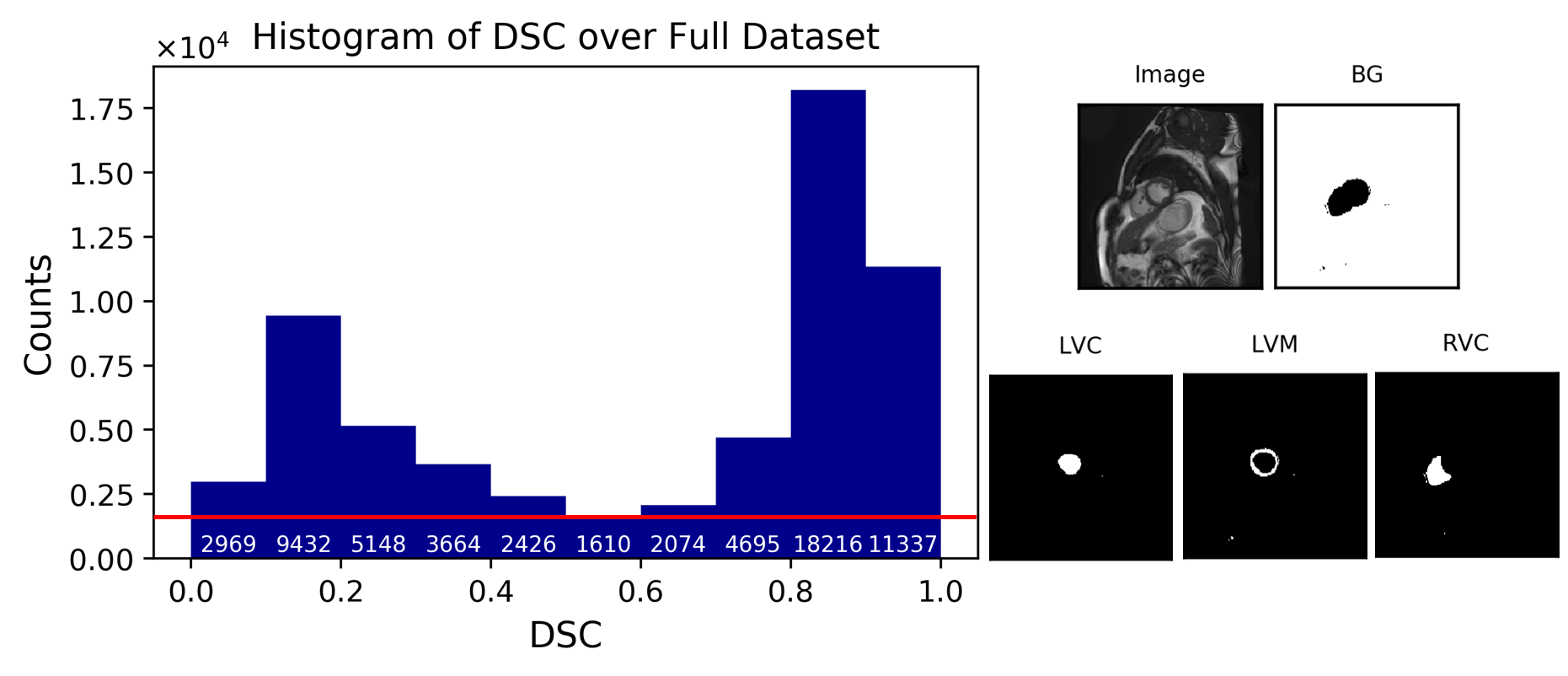}
	\caption{(left) Histogram of Dice Similarity Coefficients (DSC) for 29,292 segmentations. Range is $[0, 1]$ with 10 equally spaced bins. Red line shows minimum counts (1,610) at $\mathrm{DSC}$ in the bin $[0.5, 0.6)$ used to balance scores. (right) 5 channels of the CNNs in both experiments: the image and one-hot-encoded labelmaps for background (BG), left-ventricular cavity (LV), left-ventricular myocardium (LVM) and right-ventricular cavity (RVC).}
		\label{fig:dscdist}
\end{figure}

We use the Dice Similarity Coefficient (DSC) as a metric of quality for segmentations. It measures the overlap between a proposed segmentation and its ground truth (GT) (usually a manual reference). We aim to predict DSC for segmentations in the \textit{absence} of GT. We perform two experiments in which CNNs are trained to predict DSC. First we describe our input data and the models.

Our initial dataset consists of 4,882 3D (2D-stacks) end-diastolic (ED) cardiovascular magnetic resonance (CMR) scans from the UK Biobank (UKBB) Imaging Study\footnote{UK Biobank Resource under Application Number 2964.}. All images have a manual segmentation which is unprecedented at this scale. We take these labelmaps as reference GT. Each labelmap contains 3 classes: left-ventricular cavity (LVC), left-ventricular myocardium (LVM) and right-ventricular cavity (RVC) which are separate from the background class (BG). In this work, we also consider the segmentation as a single binary entity comprising all classes: whole-heart (WH).

A random forest (RF) of 350 trees and maximum Depth 40 is trained on 100 cardiac atlases from an in-house database and used to segment the 4,882 images at depths of 2, 4, 6, 8, 10, 15, 20, 24, 36 and 40. We calculate DSC from the GT for the 29,292 generated segmentations. The distribution is shown in Fig~\ref{fig:dscdist}. Due to the imbalance in DSC scores of this data, we choose to take a random subset of 1,610 segmentations from each DSC bin, equal to the minimum number of counts-per-bin across the distribution. Our final dataset comprises 16,100 score-balanced segmentations with reference GT.

From each segmentation we create 4 one-hot-encoded masks: masks 1 to 4 correspond to the classes BG, LVC, LVM and RVC respectively. The voxels of the $i^{th}$ mask are set at $[0,0,0,0]$ when they do not belong to the mask's class and the $i^{th}$ element set to 1 otherwise. For example, the mask for LVC is $[0,0,0,0]$ everywhere except for voxels of the LVC class which are given the value $[0,1,0,0]$. This gives the network a greater chance to learn the relationships between the voxels' classes and their locations. An example of the segmentation masks is shown in Fig~\ref{fig:dscdist}.

At training time, our data-generator re-samples the UKBB images and our segmentations to have consistent shape of $[224, 224, 8, 5]$ making our network fully 3D with 5 data channels: the image and 4 segmentation masks. The images are also normalized such that the entire dataset falls in the range $[0.0, 1.0]$.

For comparison and consistency, we choose to use the same input data and network architecture for each of our experiments. We employ a 50-layer 3D residual network written in Python with the Keras library and trained on an 11GB Nvidia GeForce GTX 1080 Ti GPU. Residual networks are advantageous as they allow the training of deeper networks by repeating smaller blocks. They benefit from skip connections that allow data to travel deeper into the network. We use the Adam optimizer with learning rate of $1e^{-5}$ and decay of 0.005. Batch sizes are kept constant at 46 samples per batch. We run validation at the end of each epoch for model-selection purposes.

\subsection*{Experiments}

Can we take advantage of a CNN's inference speed to give fast and accurate predictions of segmentation quality? This is an important question for analysis pipelines which could benefit from the increased confidence in segmentation quality without compromising processing time. To answer this question we conduct the following experiments.

\subsubsection*{Experiment 1: Directly predicting DSC.} Is it possible to directly predict the quality of a segmentation given only the image-segmentation pair? In this experiment we calculate, per class, the DSC between our segmentations and the GT. These are used as training labels. We have 5 nodes in the final layer of the network where the output $X$ is $\lbrace X \in \mathbb{R}^{5} \ | X \in [0.0,1.0] \rbrace$. This vector represents the DSC per class including background and whole-heart. We use mean-squared-error loss and report mean-absolute-error between the output and GT DSC. We split our data 80:10:10 giving 12,880 training samples and 1,610 samples each for validation and testing. Performing this experiment is costly as it requires a large manually-labeled dataset which is not readily available in practice.

\subsubsection*{Experiment 2: Predicting RCA scores.} Considering the promising results of the RCA framework \cite{Valindriaa,Robinson2017} in accurately predicting the quality of segmentations in the absence of large labeled datasets, can we use the predictions from RCA as training data to allow a network to give comparatively accurate predictions on a test-set? In this experiment, we perform RCA on all 16,100 segmentations. To ensure that we train on balanced scores, we again perform histogram binning on the RCA scores and take equal numbers from each class. We finish with a total of 5363 samples split into training, validation and test sets of 4787, 228 and 228 respectively. The predictions per-class are used as labels during training. Similar to Experiment 1, we obtain a single predicted DSC output for each class using the same network and hyper-parameters, but without the need for the large, often-unobtainable manually-labeled training set.

%% file: sections/results.tex

\section{Results}
\label{sec:results}

\begin{figure}[tb]
	\includegraphics[width=\linewidth]{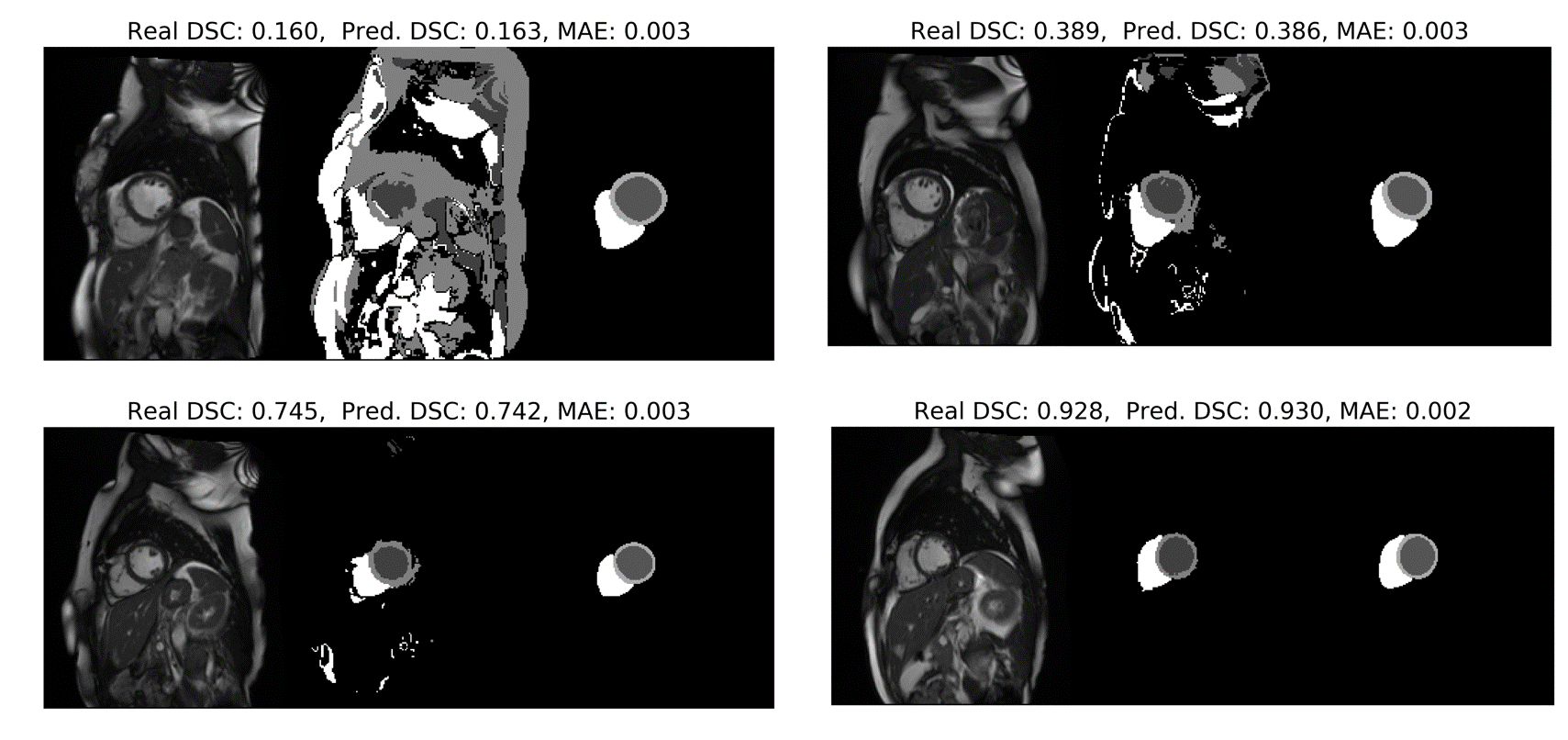}
      \caption{Examples showing excellent prediction of Dice Similarity Coefficient (DSC) in Experiment 1. Quality increases from top-left to bottom-right. Each panel shows (left to right) the image, test-segmentation and reference GT.\label{fig:examples}}	\end{figure}

\begin{figure}[tb]
	\includegraphics[width=\linewidth]{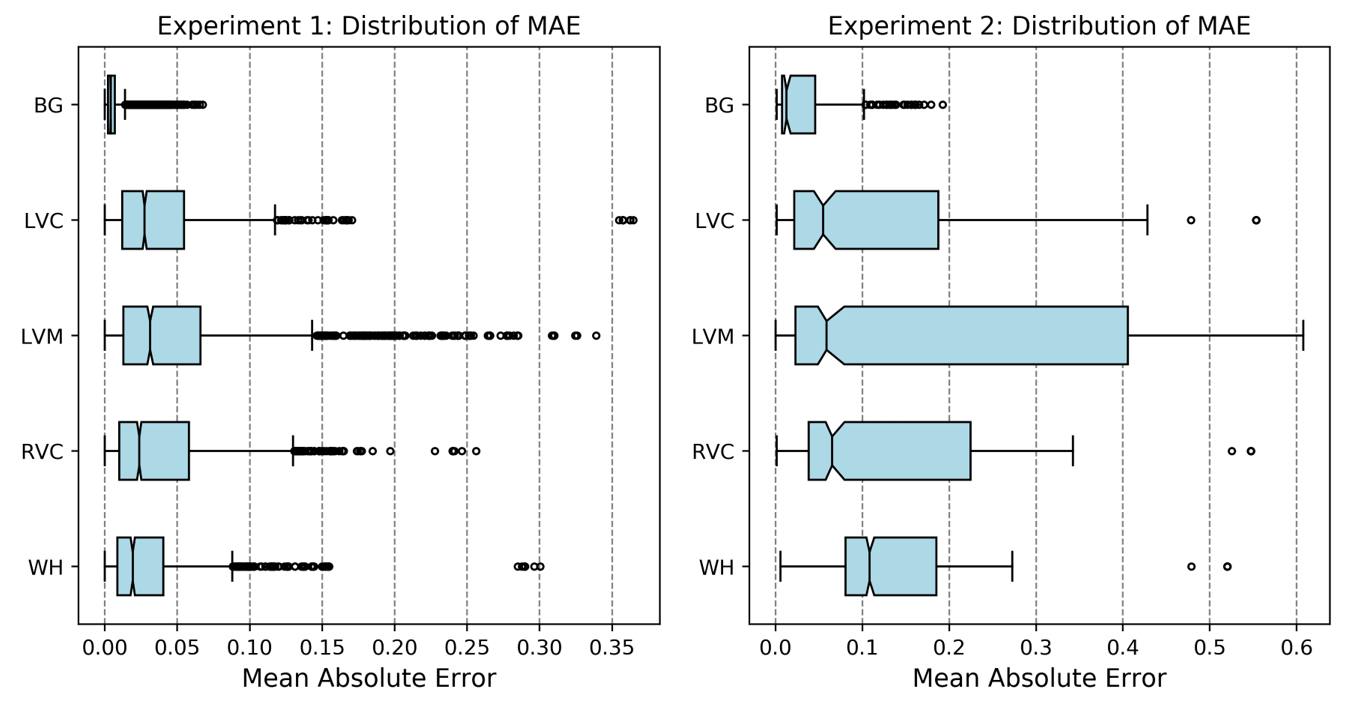}
      \caption{Distribution of the mean absolute errors (MAE) for Experiments 1 (left) and 2 (right). Results are shown for each class: background (BG), left-ventricular cavity (LV), left-ventricular myocardium (LVM), right-ventricular cavity (RVC) and for the whole-heart (WH). \label{fig:maeboxplots}}	\end{figure}

Results from Experiment 1 are shown in Table~\ref{tab:results1}. We report mean absolute error (MAE) and standard deviations per class between reference GT and predicted DSC. Our results show that our network can directly predict whole-heart DSC from the image-segmentation pair with MAE of 0.03 (SD = 0.04). We see similar performance on individual classes. Table~\ref{tab:results1} also shows MAE over the top and bottom halves of the GT DSC range. This suggests that the MAE is equally distributed over poor and good quality segmentations. For WH we report 72\% of the data have MAE less than 0.05 with outliers ($\mathrm{DSC} \geq 0.12$) comprising only 6\% of the data. Distributions of the MAEs for each class can be seen in Fig~\ref{fig:maeboxplots}. Examples of good and poor quality segmentations are shown in Fig~\ref{fig:examples} with their GT and predictions. Results show excellent true (TPR) and false-positive rates (FPR) on a whole-heart binary classification task with DSC threshold of 0.70. The reported accuracy of 97\% is better than the 95\% reported with RCA in \cite{Robinson2017}.

Our results for Experiment 2 are recorded in Table~\ref{tab:results1}. It is expected that direct predictions of DSC from the RCA labels are less accurate than in Experiment 1. The reasoning is two-fold: first, the RCA labels are themselves \textit{predictions} and retain inherent uncertainty and second, the training set here is much smaller than in Experiment 1. However, we report MAE of 0.14 (SD = 0.09) for the WH case and 91\% accuracy on the binary classification task. Distributions of the MAEs are shown in Fig~\ref{fig:maeboxplots}. LVM has a greater variance in MAE which is in line with previous results using RCA \cite{Robinson2017}. Thus, the network would be a valuable addition to an analysis pipeline where operators can be informed of likely poor-quality segmentations, along with some confidence interval, in real-time.

On average, the inference time for each network was of the order 600 ms on CPU and 40 ms on GPU. This is over 10,000 times faster than with RCA (660 seconds) whilst maintaining good accuracy. In an automated image analysis pipeline, this method would deliver excellent performance at high-speed and at large-scale. When paired with a real-time segmentation method it would be possible provide real-time feedback during image acquisition whether an acquired image is of sufficient quality for the downstream segmentation task.

\begin{table}[tbh!]
  \caption{For Experiments 1 and 2, Mean absolute error (MAE) for poor ($\mathrm{DSC} < 0.5$) and good ($\mathrm{DSC} \geq 0.5$) quality segmentations over individual classes and whole-heart (WH). Standard deviations in brackets. (right) Statistics from binary classification (threshold $\mathrm{DSC} =0.7$ \cite{Robinson2017}): True (TRP) and false-positive (FPR) rates over full DSC range with classification accuracy (Acc). \label{tab:results1}}
  \centering
  \scriptsize
    \centering
\resizebox{\columnwidth}{!}{
    \begin{minipage}[t]{\linewidth}
  \begin{tabular}{r |ccc |ccc|}

   \multicolumn{1}{c}{}&\multicolumn{6}{c}{\textbf{Mean Absolute Error (MAE)}} \\
  \cline{2-7}
    \multicolumn{1}{c}{} & \multicolumn{3}{|c}{\textbf{Experiment 1}} & \multicolumn{3}{|c|}{\textbf{Experiment 2}} \\
  \cline{2-4} \cline{5-7}
   			\multicolumn{1}{c|}{}	& 0 $\leq$ DSC $\leq$ 1 		& $\mathbf{DSC < 0.5}$ 		& $\mathbf{DSC \geq 0.5}$ 	& \textbf{0 $\leq$ DSC $\leq$ 1} 		& $\mathbf{DSC < 0.5}$ 		& $\mathbf{DSC \geq 0.5}$\\
 \textbf{Class}            & $n=1,610$ 	& $n=817$ 		& $n=793$ 	& $n=288$ 		&$n=160$ 		& $n=128$\\ 	\hline 		
  BG 			& 0.008 (0.011)		& 0.012 (0.014)			& 0.004 (0.002) & 0.034 (0.042)	&0.048 (0.046)	&0.074 (0.002)	\\
  LV 			& 0.038 (0.040)		& 0.025 (0.024)			& 0.053 (0.047)	& 0.120 (0.128)	&0.069 (0.125)	&0.213 (0.065)\\
  LVM 			& 0.055 (0.064)		& 0.027 (0.027)			& 0.083 (0.078)	& 0.191 (0.218)	&0.042 (0.041)	&0.473 (0.111)\\	
  RVC	 		& 0.039 (0.041)		& 0.021 (0.020)			& 0.058 (0.047)	& 0.127 (0.126)	&0.076 (0.109)	&0.223 (0.098)\\	
  \textbf{WH} 	&\textbf{0.031 (0.035)}	& \textbf{0.018 (0.018)}	& \textbf{0.043 (0.043)}	& \textbf{0.139 (0.091)}	&\textbf{0.112 (0.093)}	&\textbf{0.188 (0.060)}\\	 \hline 
 \multicolumn{1}{c}{} & \multicolumn{1}{|c}{TPR 0.975}	&	 \multicolumn{1}{c}{FPR 0.060}	& \multicolumn{1}{c|}{\textbf{Acc.} \textbf{0.965}} & \multicolumn{1}{c}{TPR 0.879}	&	\multicolumn{1}{c}{FPR 0.000}	& \multicolumn{1}{c|}{\textbf{Acc.} \textbf{0.906}} \\ \cline{2-7}
  \end{tabular}
  \end{minipage}
}
\end{table}

%% file: sections/conclusion.tex

\section{Conclusion}
\label{sec:conclusion}

Ensuring the quality of a automatically generated segmentation in a deployed image analysis pipeline in real-time is challenging. We have shown that we can employ Convolutional Neural Networks to tackle this problem with great computational efficient and with good accuracy.

We recognize that our networks are prone to learning features specific to assessing the quality of Random Forest segmentations. We can build on this by training the network with segmentations generated from an ensemble of methods. However, we must reiterate that the purpose of the framework in this study is to give an indication of the \textit{predicted quality} and not a direct one-to-one mapping to the reference DSC. Currently, these networks will correctly predict whether a segmentation is `good' or `poor' on some threshold, but will not confidently distinguish between two segmentations of similar quality.

Our trained CNNs are insensitive to small regional or boundary differences in labelmaps which are of good quality. Thus they cannot be used to assess quality of a segmentation at fine-scale. Again, this may be improved by a more diverse and granular training-sets. The labels for training the network in Experiment 1 are not easily available in most cases. However, by performing RCA, one can automatically obtain training labels for the network in Experiment 2 and this could be applied to segmentations generated with other algorithms. The cost of using data obtained with RCA is an increase in MAE. This is reasonable compared to the effort required to obtain a large, manually-labeled dataset.